\PassOptionsToPackage{table}{xcolor}

\documentclass[11pt]{article}

\usepackage[preprint]{acl}
\usepackage{times}
\usepackage{latexsym}
\usepackage{booktabs}
\usepackage{multirow}
\usepackage{amssymb}
\usepackage{algorithm}

\usepackage{graphicx}
\usepackage[export]{adjustbox}

\usepackage{algpseudocode}
\usepackage{tabularx}
\usepackage[T1]{fontenc}
\usepackage{amsmath}
\usepackage[utf8]{inputenc}

\usepackage{microtype}

\usepackage{inconsolata}

\usepackage{booktabs}

\usepackage{subcaption}

\usepackage{tabularx}
\newcolumntype{Y}{>{\centering\arraybackslash}X} 

\usepackage[utf8]{inputenc}
\usepackage[T1]{fontenc}
\usepackage{geometry}
\usepackage[table]{xcolor}

\usepackage[most]{tcolorbox}

\definecolor{boxBorderColor}{RGB}{120, 120, 220} 
\definecolor{boxTitleBgColor}{RGB}{120, 120, 220}

\newtcolorbox{PromptBox}[2][]{
    enhanced,                
    colframe=boxBorderColor, 
    colbacktitle=boxTitleBgColor, 
    coltitle=white,         
    colback=white,          
    fonttitle=\bfseries\large, 
    arc=3mm,                
    boxrule=1.2pt,           
    titlerule=0pt,          
    title={#2},              
    #1                     
}

\usepackage{graphicx}

\newcommand{\ours}{\textbf{\texttt{SARE}}}

\title{Beyond Superficial Unlearning: Sharpness-Aware Robust Erasure \\ of Hallucinations in Multimodal LLMs}

\author{
\textbf{Xianya Fang}\textsuperscript{$\clubsuit$}\footnotemark[1]~,  
\textbf{Feiyang Ren}\textsuperscript{$\clubsuit$}\thanks{\;Equal Contribution.}~,
\textbf{Xiang Chen}\textsuperscript{$\clubsuit$}\thanks{\;Corresponding Author.},
\textbf{Yu Tian}\textsuperscript{$\spadesuit$},\\ 
\textbf{Zhen Bi}\textsuperscript{$\diamondsuit$},
\textbf{Haiyang Yu}\textsuperscript{$\heartsuit$}\textsuperscript{$\blacksquare$}, 
\textbf{Sheng-Jun Huang}\textsuperscript{$\clubsuit$}
\\
  \textsuperscript{$\clubsuit$}College of Computer Science and Technology, \\Nanjing University of Aeronautics and Astronautics \\
  \textsuperscript{$\spadesuit$}Institute for AI, Tsinghua University 
  \textsuperscript{$\diamondsuit$}Huzhou University \\
  \textsuperscript{$\heartsuit$}Institute of Dataspace, Hefei Comprehensive National Science Center \\ 
  \textsuperscript{$\blacksquare$}University of Science and Technology of China
 \\
    {  \texttt{\{xyfang,xiang\_chen\}@nuaa.edu.cn}}\\
}

\begin{document}
\maketitle
\begin{abstract}
Multimodal LLMs are powerful but prone to object hallucinations, which describe non-existent entities and harm reliability. While recent unlearning methods attempt to mitigate this, we identify a critical flaw: structural fragility. We empirically demonstrate that standard erasure achieves only superficial suppression, trapping the model in sharp minima where hallucinations catastrophically resurge after lightweight relearning. To ensure geometric stability, we propose SARE, which casts unlearning as a targeted min–max optimization problem and uses a Targeted-SAM mechanism to explicitly flatten the loss landscape around hallucinated concepts. By suppressing hallucinations under simulated worst-case parameter perturbations, our framework ensures robust removal stable against weight shifts. Extensive experiments demonstrate that SARE significantly outperforms baselines in erasure efficacy while preserving general generation quality. Crucially, it maintains persistent hallucination suppression against relearning and parameter updates, validating the effectiveness of geometric stabilization. Our code is available at \url{https://github.com/user007-hash/SARE}.
\end{abstract}

\begin{figure}[t!]
    \centering
    \begin{subfigure}[b]{1.0\linewidth} 
        \centering
        \includegraphics[width=0.9\linewidth]{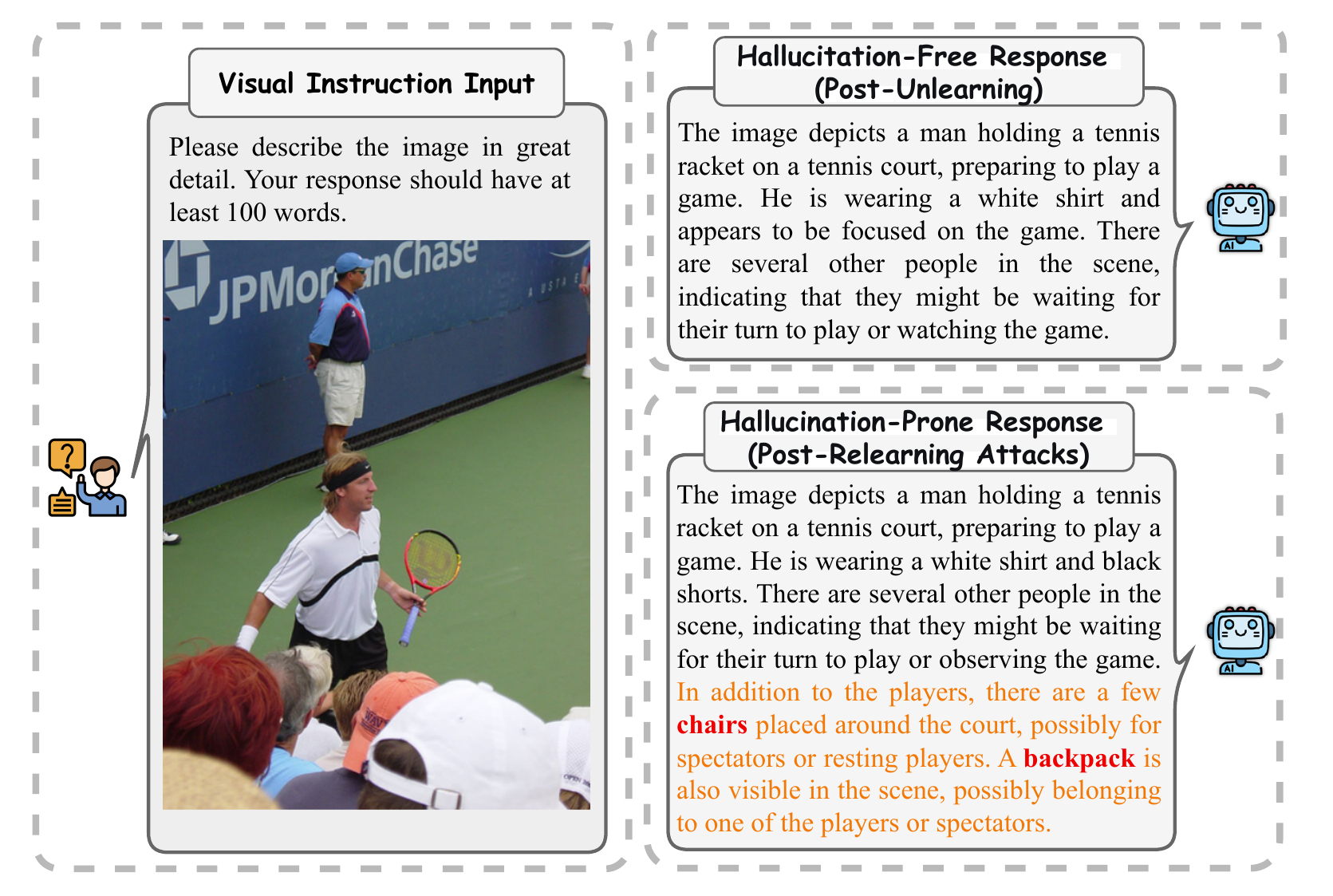}
        \caption{Vulnerability of existing unlearning methods in MLLMs}
        \label{fig:insight-a}
    \end{subfigure}
    
    \vspace{1em} 
    
    \begin{subfigure}[b]{1.0\linewidth} 
        \centering
        \includegraphics[width=0.9\linewidth]{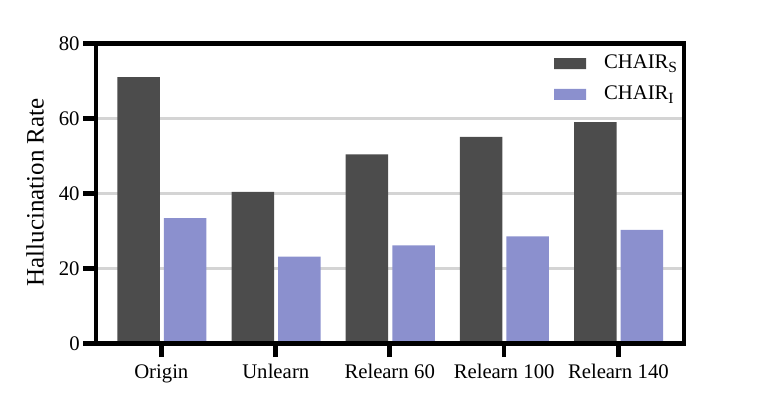}
        \caption{Hallucination Rates under Relearning}
        \label{fig:insight-b}
    \end{subfigure}
    \vspace{-1pt}
    \caption{\textbf{The vulnerability of unlearned MLLMs against relearning attacks.} (a) Lightweight relearning can easily reactivate suppressed hallucinations. (b) The hallucination rate of EFUF exhibits a rapid resurgence as the number of relearning samples increases.}
    \label{fig:insight}
\end{figure}

\section{Introduction}
\label{sec:intro}

Multimodal Large Language Models (MLLMs) have reshaped the landscape of vision-language tasks, demonstrating exceptional proficiency in image captioning~\cite{LaiS0CZZHTGGCY25,Li2023BLIP2BL,Captionanything} and visual question answering~\cite{FlowVQA,VQAI}. However, this prowess is shadowed by the persistent issue of hallucination~\cite{Liu2024ASO,ChenFactCHDBF,BitenGK22,POPE}, where generated text conflicts with visual evidence. 
Such unfaithful outputs fundamentally undermine the trustworthiness of MLLMs in real-world applications, necessitating effective mitigation strategies.

Existing countermeasures typically fall into two categories: \emph{training-stage alignment}~\cite{GunjalYB24,SunSCLLSGGWYKD24,YuYZHHCHL0024,Zhao2024AlignGPTML} and \emph{inference-stage intervention}~\cite{YinFZXWSSLSC24,ShareGPT4VIL}. While effective, the former incurs prohibitive data annotation and retraining costs, whereas the latter often imposes significant inference latency. Consequently, \textbf{machine unlearning} has emerged as a compelling, resource-efficient paradigm~\cite{Cao2015TowardsMS,Ullah2021MachineUV}. By selectively ``forgetting'' specific hallucination patterns without full retraining, methods like EFUF~\cite{EFUF} promise a balance between efficiency and safety.

However, we argue that current unlearning approaches for MLLMs remain \textit{superficial}. Our empirical analysis reveals a structural fragility: models subjected to standard unlearning exhibit a rapid resurgence of hallucinations when subjected to relearning attacks~\cite{Lynch2024EightMT,Deeb2024DoUM}, which involve exposure to a negligible amount of the original hallucination-inducing data. In practical deployment, this vulnerability is highly critical. MLLMs frequently undergo incremental fine-tuning to adapt to specialized tasks. Since real-world data distributions inevitably overlap with pre-training data, even standard non-malicious updates can inadvertently act as unintended relearning, shifting parameters and reactivating suppressed errors. As shown in Figure~\ref{fig:insight-a}, unlearned models initially produce hallucination-free captions, but quickly revert to hallucination-prone behaviour after only tens of relearning samples. Quantitatively, Figure~\ref{fig:insight-b} shows that EFUF's hallucination rate rises sharply with more relearning samples, regressing to the pre-unlearning baseline.

From an optimization perspective, we hypothesize that this fragility arises because model parameters are trapped in a \textit{sharp minimum} of the hallucination loss landscape. In such a precarious configuration, hallucination suppression is highly sensitive, as even minor parameter changes introduced by relearning can quickly move the model back into a hallucination-prone region. This suggests that the unwanted knowledge is not truly erased but only suppressed at a sharp local basin, leaving the model vulnerable to regression under subsequent training.

To address this, we propose \textbf{S}harpness-\textbf{A}ware \textbf{R}obust \textbf{E}rasure of Hallucinations in Multimodal LLMs (\textbf{\ours}), a novel framework that enforces robust erasure through geometric regularization. Drawing inspiration from Sharpness-Aware Minimization (SAM)~\cite{SAM}, \ours~reformulates the unlearning objective as a targeted min-max problem. Instead of simply minimizing the unlearning loss, \ours~ simulates a worst-case attack by identifying the weight perturbation most likely to reactivate hallucinations, and minimizes the loss under this adversarial condition. This process effectively flattens the loss landscape around the unlearned state, ensuring that the erasure of hallucinations is stable and invariant to small weight shifts. Crucially, {\ours}~ retains the data-efficient curation pipeline of existing methods while fundamentally upgrading the optimization mechanism. Our contributions are summarized as follows:
\begin{itemize}
    \item To the best of our knowledge, we are the first to reveal the robustness gap in MLLM hallucination unlearning, collapsing rapidly under lightweight relearning attacks.
    \item We introduce \textbf{\ours}, a sharpness-aware framework optimizing for flat minima via Targeted-SAM to ensure deep, durable hallucination erasure while preserving general capabilities.
    \item \ours~ achieves persistent hallucination erasure that resists regression. Extensive experiments verify its stability against relearning, fine-tuning, and adversarial prompting.
\end{itemize}

\section{Related Work}
\label{sec:related_work}

\subsection{Hallucination Mitigation of MLLMs}

Hallucination in MLLMs~\cite{ZhuJCXY025,POPE} refers to cross-modal misalignment where textual outputs contradict visual evidence~\cite{Liu2024ASO,ChenFactCHDBF}. While this phenomenon encompasses broader categories such as temporal inconsistencies or complex logical fallacies, the most prevalent and detrimental subtype remains object hallucination, which describes non-existent items~\cite{Liu2024ASO,POPE,BitenGK22}. Addressing this foundational issue is critical for ensuring model safety in real-world applications~\cite{Captionanything,ZhaoWODWH25,HuangDZ0H0L0Y24,Zhang2023SirensSI}.

Existing mitigation strategies for MLLMs can be categorized based on their functional roles and specific stages in the model development life cycle, rather than their underlying optimization mechanisms. Training-stage methods, such as fine-tuning on specialized datasets~\cite{YouZGDZWCCY24,GunjalYB24} or applying advanced alignment objectives like RLHF~\cite{SunSCLLSGGWYKD24,YuYZHHCHL0024,Zhao2024AlignGPTML}, refer to the foundational alignment phase where the model learns general instructions using large-scale datasets. Inference-stage methods, including post-hoc revision~\cite{YinFZXWSSLSC24} and training-free decoding~\cite{ShareGPT4VIL,JiLFYSXIBMF23}, intervene during the decoding process to rectify outputs without modifying any parameters, which significantly increases latency. In contrast, machine unlearning serves as a distinct post-training strategy. It directly updates the parameters of finalized models, offering an efficient middle ground between the heavy burden of foundational alignment and the latency overhead of inference-stage interventions.

\subsection{Unlearning and Its Adversarial Robustness in LLMs}

 Machine unlearning refers to a technique designed to enable models to selectively erase specific data or behaviors while preserving general utility, serving as an efficient alternative to full retraining~\cite{Cao2015TowardsMS,Ullah2021MachineUV}. Common unlearning techniques primarily employ methods like gradient ascent~\cite{JangYYCLLS23} or KL-divergence constraints to balance these objectives~\cite{YuYZHHCHL0024}.
However, existing methods are vulnerable to adversarial attacks. Of particular concern are relearning attacks, where fine-tuning the unlearned model with even a small amount of the original “forgotten” data can rapidly restore the undesired knowledge and behaviors~\cite{Lynch2024EightMT,Deeb2024DoUM}. Pioneering work like EFUF has introduced unlearning to multimodal hallucination mitigation, but inherits these robustness limitations. In our research, we address this by developing a robust unlearning framework for MLLMs to defend against such attacks.

\subsection{Sharpness Awareness Minimization}

Sharpness-Aware Minimization (SAM) enhances model generalization by guiding the training towards parameters lying in neighborhoods with uniformly low loss, thereby promoting a flat loss landscape~\cite{SAM}. Formulated as a min-max optimization problem, SAM and its variants (e.g., ASAM~\cite{ASAM}, ESAM~\cite{ESAM}) explicitly pursue flat minima to improve generalization~\cite{BartlettLB23,Ujvary2022RethinkingSM}. Beyond generalization, this smoothness is linked to robustness: SAM’s mechanism of optimizing against worst-case parameter perturbations has proven effective in adversarial training to defend against input-level attacks~\cite{Wei2023SharpnessAwareMA,Liu2024RevisitingWH}. Recent theoretical insights further reveal a duality between SAM and Adversarial Training (AT) and demonstrate that SAM learns more robust feature representations~\cite{Zhang2024OnTD}. Motivated by these principles, we integrate SAM's smoothness optimization to develop a robust unlearning framework for MLLMs specifically designed to withstand relearning attacks.

\section{Methodology}
\label{sec:method}

\subsection{Why Standard Unlearning Fails to Erase Hallucinations?}
\label{subsec:insights}

Standard unlearning typically employs a multi-objective optimization strategy: minimizing the likelihood of hallucinated samples while maintaining performance on normal data. Formally, the baseline objective is defined as:
\begin{equation}
\label{eq:base_loss}
\mathcal{L}_{base}(\theta_{\phi}) = \mathcal{L}_{pos} + \lambda_1 \mathcal{L}_{neg} + \lambda_2 \mathcal{L}_{sent},
\end{equation}
where $\mathcal{L}_{neg}$ suppresses spurious hallucination correlations, while $\mathcal{L}_{pos}$ and $\mathcal{L}_{sent}$ preserve visual grounding and linguistic capabilities.

However, directly minimizing $\mathcal{L}_{base}$ traps the model in a \textit{sharp minimum}. As shown in Figure \ref{fig:insight}, this geometric characteristic implies extreme sensitivity to parameter changes: while the model successfully suppresses hallucinations at the exact optimal weights, adding even a slight perturbation causes the hallucination rate to spike drastically. This indicates that the erasure is structurally fragile, the unlearning effect is strictly confined to a precise point and cannot withstand the weight shifts inherent in model deployment or subsequent tuning. 
To bridge this gap, we draw inspiration from the adversary-defense game perspective~\cite{SAM1}. We hypothesize that true hallucination mitigation demands geometric stability—a flat loss landscape where the erasure remains effective even when parameters drift. Thus, we frame unlearning as a targeted min-max robust optimization problem.

\begin{figure*}[t]
    \centering
    \includegraphics[width=1.0\textwidth]{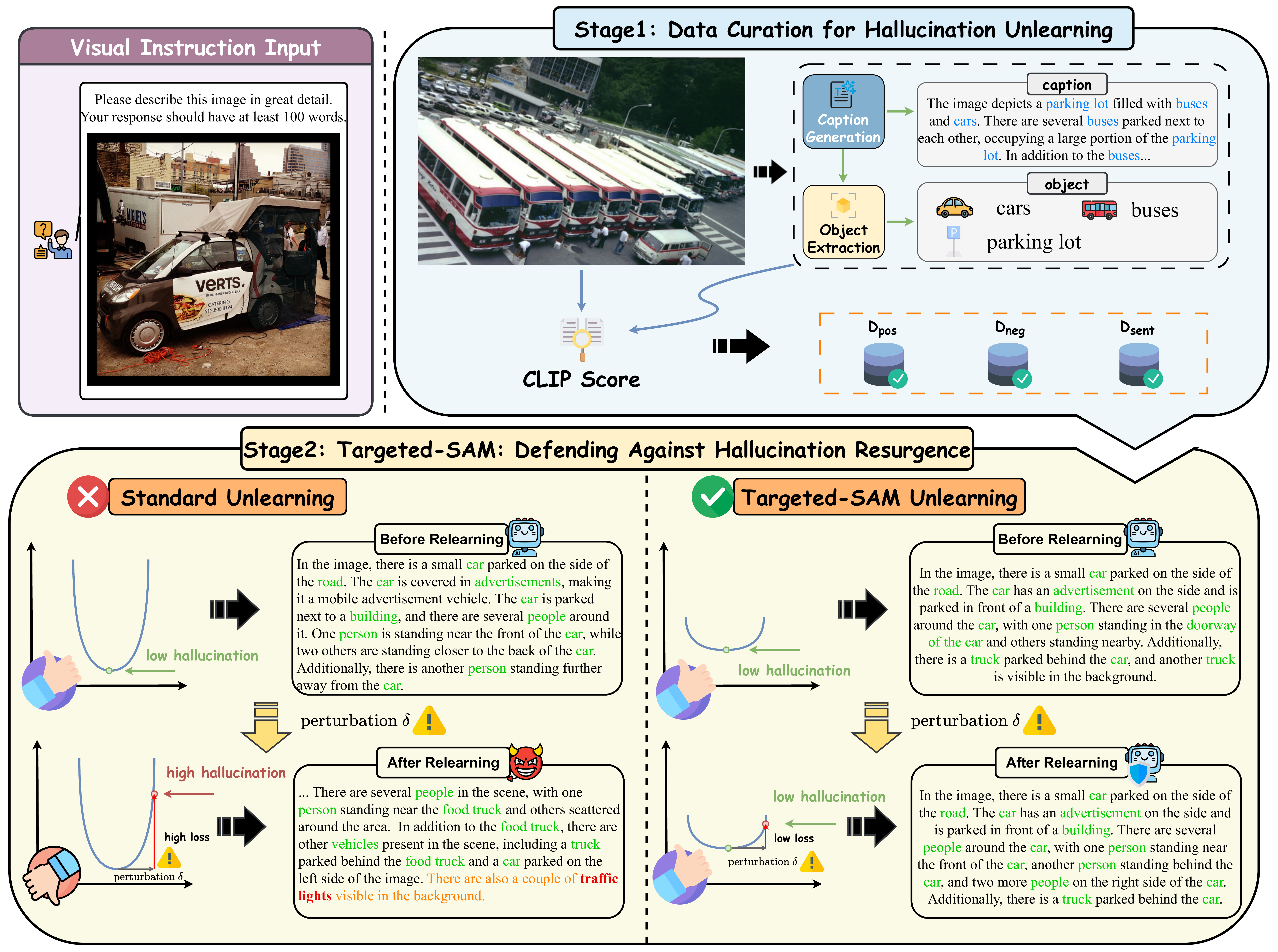} 
    \vspace{0.5em}
    \caption{Overview of the \ours~ framework. The top-right panel illustrates Stage 1, where an automated pipeline curates training subsets ($D_{neg}, D_{pos}, D_{sent}$). The bottom panel depicts Stage 2, contrasting the fragile sharp minima of standard unlearning (left) with the robust flat loss landscape of \ours~ (right).}
    \label{fig:demo}
\end{figure*}

\subsection{SARE: A Framework for Robust Hallucination Erasure}

Based on these insights, we propose \textbf{\ours}, a robust unlearning framework designed to harmonize data efficiency with optimization stability. As illustrated in Figure \ref{fig:demo}, our approach unfolds in two stages:
The first stage is \textbf{Data Curation for Hallucination Unlearning}. 
We adopt an automated pipeline established by EFUF~\cite{EFUF} to curate subsets: the negative subsentence dataset ($D_{neg}$) containing object hallucinations for erasure, the positive subsentence dataset ($D_{pos}$) for retaining visual grounding, and the sentence-level dataset ($D_{sent}$) to preserve linguistic coherence.

The second stage is \textbf{Targeted Sharpness Tuning}. To address the vulnerability of sharp minimum, we implement a Targeted-SAM mechanism. Rather than standard gradient descent, we first simulate a worst-case attack by maximizing the likelihood of hallucination relapse on $D_{neg}$, and then minimize the joint objective to suppress hallucinations and preserve capabilities on $D_{pos}$ and $D_{sent}$ under this worst-case interference. This enforces a flat loss landscape, ensuring that the erasure of hallucinations remains stable against weight shifts.

\subsection{Data Curation for Hallucination Unlearning}
\label{subsec:data_loss}

To circumvent the prohibitive costs of manual annotation, we directly leverage the automated data curation pipeline established by EFUF~\cite{EFUF}. We employ CLIP-based alignment scores as a reliable proxy for visual grounding: high scores indicate accurately recognized objects, while low scores identify hallucinated content.
Formally, for each object $o$ identified in a response, the training unit is denoted as $u_o = (v, \text{pre}(o), \text{cur}(o))$, where $v$ is the visual input, $\text{cur}(o)$ is the subsentence describing object $o$, and $\text{pre}(o)$ captures the preceding context. 
We adopt EFUF's determined thresholds $T_0$ for high-confidence grounding and $T_1$ for hallucinated content to categorize object scores $S(o)$ into visual anchors ($D_{pos}$) and targets ($D_{neg}$):
\begin{equation}
\begin{aligned}
    D_{pos} &= \{ u_o \mid S(o) > T_0 \}, \\
    D_{neg} &= \{ u_o \mid S(o) < T_1 \}.
\end{aligned}
\end{equation}

Additionally, to prevent the unlearning process from degrading the model's fluency or instruction-following abilities, we utilize the sentence-level dataset $D_{sent}$, obtained by filtering responses with an average relevance score $S(y)$ above a sentence-level reliability threshold $T_2$:
\begin{equation}
    D_{sent} = \{ (v, x, y) \mid S(y) > T_2 \},
\end{equation}
comprising the image $v$, prompt $x$, and the reliable response $y$.

With these curated subsets, we explicitly define the loss components introduced in Sec.~\ref{subsec:insights}. Using the standard fine-tuning objective $\mathcal{L}_{ft}$, we minimize the loss on positive anchors and general instructions to preserve grounding, while inverting the loss on negative targets to penalize hallucination:
\begin{equation}
\begin{aligned}
    \mathcal{L}_{pos}(\theta_{\phi}) &= \mathcal{L}_{ft}(v, \text{pre}(o), \text{cur}(o)), \quad u_o \in D_{pos}, \\
    \mathcal{L}_{neg}(\theta_{\phi}) &= -\mathcal{L}_{ft}(v, \text{pre}(o), \text{cur}(o)), \quad u_o \in D_{neg}, \\
    \mathcal{L}_{sent}(\theta_{\phi}) &= \mathcal{L}_{ft}(v, x, y), \quad (v, x, y) \in D_{sent}.
\end{aligned}
\end{equation}
These objectives provide the foundational signals for our subsequent Targeted-SAM.

\subsection{Targeted-SAM: Defending Against Hallucination Resurgence}

\paragraph{Min-Max Formulation and Sharpness Regularization.}
Formulating the robust unlearning process as a min-max problem, we design a targeted adversarial attack to simulate the worst-case scenario where hallucinations are most likely to resurface. Specifically, the inner maximization seeks a parameter perturbation $\epsilon$ that maximizes the probability of generating hallucinated objects ($\mathcal{L}_{neg}$), effectively exposing the model's most vulnerable geometric direction. The outer minimization then suppresses hallucinations under this worst-case interference while anchoring valid capabilities on the original weights:
\begin{equation}
\label{eq:minmax_obj}
\begin{split}
    \min_{\theta_{\phi}} \Bigg( & \underbrace{\max_{\|\epsilon\|_2 \le \rho} \lambda_1 \mathcal{L}_{neg}(\theta_{\phi} + \epsilon)}_{\text{Targeted Perturbation}} \\
    & + \underbrace{\mathcal{L}_{pos}(\theta_{\phi}) + \lambda_2 \mathcal{L}_{sent}(\theta_{\phi})}_{\text{Capability Preservation}} \Bigg),
\end{split}
\end{equation}
where $\rho > 0$ is a small hyperparameter that controls the neighborhood radius of the perturbation.

To efficiently solve the inner maximization in Eq.~\eqref{eq:minmax_obj}, we approximate the optimal perturbation $\epsilon^*$ using a first-order Taylor expansion. This simplifies the optimization to aligning with the gradient direction of the hallucination loss, thereby maximizing the likelihood of generating non-existent objects:
\begin{equation}
\begin{split}
    \epsilon^* &= \mathop{\arg\max}_{\|\epsilon\|_2 \le \rho} \mathcal{L}_{neg}(\theta_{\phi} + \epsilon) \\
    &\approx \rho \frac{\nabla_{\theta_{\phi}} \mathcal{L}_{neg}(\theta_{\phi})}{\|\nabla_{\theta_{\phi}} \mathcal{L}_{neg}(\theta_{\phi})\|_2}.
\end{split}
\end{equation}
This derivation reveals that the optimal perturbation $\epsilon^*$ serves as a directional indicator, pointing directly towards the region of highest vulnerability in the parameter space where the unlearned hallucinations are most liable to recur.

Substituting $\epsilon^*$ back into Eq.~\eqref{eq:minmax_obj} transforms the inner maximization into a penalized objective:
\begin{equation}
\label{eq:taylor_expansion}
    \mathcal{L}_{neg}(\theta_{\phi} + \epsilon^*) \approx \mathcal{L}_{neg}(\theta_{\phi}) + \rho \|\nabla_{\theta_{\phi}} \mathcal{L}_{neg}(\theta_{\phi})\|_2.
\end{equation}
Here, the gradient norm term $\rho \|\nabla_{\theta_{\phi}} \mathcal{L}_{neg}\|_2$ functions as a sharpness regularizer. Minimizing this term explicitly flattens the loss landscape, reducing the model's sensitivity to perturbations that would otherwise trigger hallucination relapse.
However, directly minimizing Eq.~\eqref{eq:taylor_expansion} is computationally prohibitive as differentiating the norm term $\|\nabla \mathcal{L}_{neg}\|_2$ requires the Hessian matrix (second-order derivatives).

\paragraph{Efficient Gradient Approximation and Final Update.}
To minimize Eq.~\eqref{eq:taylor_expansion} efficiently without incurring the cost of Hessian computation, we leverage the first-order approximation strategy from SAM~\cite{SAM}. SAM demonstrates that the gradient of the sharpness-regularized objective can be effectively approximated by the gradient computed at the perturbed state $\theta_{\phi} + \epsilon^*$ (see Appendix~\ref{app:sam_derivation} for the detailed derivation).
Consequently, the final aggregated gradient $g_{final}$ integrates two complementary signals: a robust suppression term computed at the worst-case perturbed state to penalize hallucination sensitivity, and standard preservation terms computed at the current parameter state to maintain general capabilities:
\begin{equation}
\begin{split}
    g_{final} = \, &  \lambda_1 \nabla_{\theta_{\phi}} \mathcal{L}_{neg}(\theta_{\phi} + \epsilon^*) \\
    & + \nabla_{\theta_{\phi}} \mathcal{L}_{pos}(\theta_{\phi}) + \lambda_2 \nabla_{\theta_{\phi}} \mathcal{L}_{sent}(\theta_{\phi}).
\end{split}
\end{equation}
This optimization ensures that the erasure of hallucinations is not merely a superficial masking at a sharp point, but a robust removal stable within a neighborhood of parameters, thereby effectively defending against relearning attacks.

\section{Experiments}
\label{sec:experiments}

\begin{table*}[t]
\centering
\resizebox{\textwidth}{!}{
\begin{tabular}{l l l | ccccc | ccccc}
\toprule
\multirow{2}{*}{\textbf{Base Model}} & \multirow{2}{*}{\textbf{Method}} & \multirow{2}{*}{\textbf{Setting}} & \multicolumn{5}{c|}{\textbf{Hallucination Rate}} & \multicolumn{5}{c}{\textbf{Generation Quality}} \\
& & & $\text{Chair}_S \downarrow$ & $\text{Chair}_I \downarrow$ & $\text{Human}_S \downarrow$ & $\text{Human}_I \downarrow$ & POPE $\uparrow$ & Bleu1 $\uparrow$ & Bleu2 $\uparrow$ & Bleu4 $\uparrow$ & Info. $\uparrow$ & ppl. $\downarrow$ \\ 
\midrule
\multirow{13}{*}{\textbf{LLaVA-v1.5-7B}} & None & Baseline & 49.4 & 22.1 & 42.0 & 14.7 & 85.3 & 43.3 & 29.2 & 15.4 & \underline{93.7} & 0.139 \\
\cmidrule{2-13}
& \multirow{6}{*}{EFUF} & Origin & 41.8 & 18.3 & 24.0 & 7.7 & 85.9 & 47.1 & 32.7 & 18.2 & 93.5 & 0.113 \\
& & w/ Relearn 60 & 45.8 & 19.7 & 26.0 & 10.4 & 85.3 & 45.8 & 31.5 & 17.3 & 93.5 & 0.111 \\
& & w/ Relearn 100 & 46.5 & 19.6 & 28.0 & 11.0 & 84.3 & 45.6 & 31.3 & 17.1 & \textbf{93.9} & 0.113 \\
& & w/ Relearn 140 & 48.9 & 20.9 & 29.0 & 11.8 & 84.0 & 45.2 & 30.9 & 16.7 & \underline{93.7} & 0.114 \\
& & w/ LoRA FT & 48.2 & 20.8 & 27.0 & 10.8 & \underline{86.1} & 45.8 & 31.3 & 17.0 & 92.2 & 0.120 \\
& & w/ Adversarial & 50.5 & 23.9 & 28.0 & 11.0 & 68.5 & 44.7 & 30.5 & 16.5 & 92.6 & 0.116 \\
\cmidrule{2-13}
\rowcolor{gray!8} \cellcolor{white} & \cellcolor{white} \multirow{6}{*}{\textbf{SARE}} & Origin & \textbf{34.4} & \textbf{14.7} & \textbf{21.0} & \textbf{6.5} & \textbf{86.7} & \underline{48.0} & \underline{33.5} & \underline{18.9} & 92.1 & 0.101 \\
& & w/ Relearn 60 & \underline{34.6} & 14.9 & \textbf{21.0} & \underline{6.7} & 86.3 & 47.9 & 33.4 & 18.8 & 92.1 & \underline{0.098} \\
& & w/ Relearn 100 & 35.4 & 15.3 & \underline{22.0} & 7.9 & 86.2 & \underline{48.0} & 33.4 & 18.8 & 92.0 & \underline{0.098} \\
& & w/ Relearn 140 & \underline{34.6} & \underline{14.8} & \textbf{21.0} & 7.4 & 86.1 & \underline{48.0} & \underline{33.5} & \underline{18.9} & 92.0 & 0.099 \\
& & w/ LoRA FT & 37.9 & 17.4 & 24.0 & 9.5 & \underline{86.4} & 46.4 & 31.9 & 17.6 & 91.2 & \textbf{0.084} \\
& & w/ Adversarial & 39.9 & 17.8 & 24.0 & 9.8 & 67.7 & \textbf{48.5} & \textbf{33.9} & \textbf{19.2} & 92.4 & 0.104 \\
\midrule
\multirow{13}{*}{\textbf{mPLUG-Owl-7B}} & None & Baseline & 69.6 & 33.5 & 60.0 & 24.1 & \textbf{85.6} & 43.3 & 29.1 & 15.1 & \textbf{91.1} & 0.129 \\
\cmidrule{2-13}
& \multirow{6}{*}{EFUF} & Origin & 43.6 & 24.2 & 46.0 & 17.7 & 83.8 & 52.3 & 37.2 & 21.4 & 90.0 & 0.139 \\
& & w/ Relearn 60 & 50.5 & 26.3 & 50.0 & 18.5 & \underline{84.8} & 49.7 & 34.8 & 19.7 & 89.1 & 0.107 \\
& & w/ Relearn 100 & 55.2 & 28.6 & 52.0 & 18.9 & 82.1 & 47.7 & 33.1 & 18.4 & 89.9 & 0.107 \\
& & w/ Relearn 140 & 59.1 & 30.4 & 53.0 & 20.0 & 81.4 & 45.8 & 31.4 & 17.0 & 89.9 & \underline{0.106} \\
& & w/ LoRA FT & 58.6 & 29.7 & 52.0 & 20.3 & 82.0 & 47.5 & 32.6 & 18.0 & 89.5 & 0.107 \\
& & w/ Adversarial & 50.0 & 26.9 & 48.0 & 18.5 & 74.0 & 52.2 & 37.1 & 21.3 & 89.9 & 0.112 \\
\cmidrule{2-13}
\rowcolor{gray!8} \cellcolor{white} & \cellcolor{white} \multirow{6}{*}{\textbf{SARE}} & Origin & \textbf{37.3} & \textbf{20.8} & \textbf{38.0} & \textbf{14.2} & 82.3 & \underline{53.5} & \underline{38.2} & \underline{22.2} & 89.6 & 0.118 \\
& & w/ Relearn 60 & 42.3 & 21.8 & 40.0 & 16.3 & \underline{84.8} & 51.2 & 36.3 & 20.8 & 89.6 & 0.110 \\
& & w/ Relearn 100 & 44.9 & 23.8 & 40.0 & 16.8 & 84.3 & 50.7 & 35.8 & 20.4 & 89.0 & 0.110 \\
& & w/ Relearn 140 & 45.8 & 23.9 & 42.0 & 17.7 & 84.0 & 49.9 & 35.1 & 19.9 & 88.4 & 0.108 \\
& & w/ LoRA FT & 47.2 & 25.6 & 46.0 & 19.3 & 83.8 & 49.3 & 34.8 & 19.6 & 88.1 & \textbf{0.096} \\
& & w/ Adversarial & \underline{37.5} & \underline{21.2} & \underline{39.0} & \underline{14.7} & 71.6 & \textbf{57.2} & \textbf{41.4} & \textbf{24.6} & \underline{90.1} & 0.122 \\
\bottomrule
\end{tabular}
}
\caption{Results on Hallucination Rates (Chair, Human, POPE) and Generation Quality (Bleu, Info., ppl.) under Relearning, LoRA, and Adversarial settings. ``None'' denotes the vanilla model. Bold and underlined indicate the best and second-best performance in each column.}
\label{tab:main_exp}
\end{table*}

\subsection{Experimental Setup}

\paragraph{Dataset.}
We conduct our experiments on the MSCOCO~\cite{MSCOCO} dataset. We randomly sample 3,200 images, allocating 1,600 for validation and 1,600 for testing. For the unlearning process, the training set comprises approximately 30,000 triplets, where each training tuple consists of a negative caption, a positive caption, and a sentence-level preservation sample. Comprehensive data statistics and partitioning details are provided in Appendix~\ref{app:dataset}.

\paragraph{Metrics.}

To rigorously assess both hallucination mitigation and the preservation of model capabilities, we employ a comprehensive suite of metrics covering trustworthiness and helpfulness.
\textbf{(1) Hallucination Evaluation:} We quantify object hallucinations using CHAIR~\cite{rohrbach-etal-2018-object} for automated caption assessment, MHumanEval~\cite{YuYZHHCHL0024} for human-verified judgment, and POPE~\cite{POPE} for discriminative hallucination assessment.
\textbf{(2) Generation Quality Evaluation:} To ensure the unlearning process does not compromise linguistic quality, we calculate BLEU~\cite{BLEU} scores for textual consistency, Informativeness for the semantic coverage of visual details, and Perplexity (PPL) to monitor text fluency. Detailed definitions and implementation specifics of these metrics are provided in Appendix~\ref{app:metrics}.

\paragraph{Models and Baselines.}

To demonstrate the universality and architectural adaptability of our approach, we conduct experiments on two representative MLLMs: mPLUG-Owl-7B~\cite{mplug} and LLaVA-v1.5-7B~\cite{llava1,llava2}. For each architecture, we compare three configurations: (1) the Vanilla model; (2) the EFUF baseline; and (3) our proposed \ours{}.

\begin{table*}[t]
    \centering
    \setlength{\tabcolsep}{3.5pt}
    \resizebox{\textwidth}{!}{
        \begin{tabular}{l ccccc ccccc}
            \toprule
            \multicolumn{1}{c}{\multirow{2}{*}{\textbf{Method}}} & \multicolumn{5}{c}{\textbf{Hallucination Rate}} & \multicolumn{5}{c}{\textbf{Generation Quality}} \\
            \cmidrule(lr){2-6} \cmidrule(lr){7-11}
             & CHAIR$_s$ $\downarrow$ & CHAIR$_i$ $\downarrow$ & Hum$_s$ $\downarrow$ & Hum$_i$ $\downarrow$ & POPE $\uparrow$ & Bleu-1 $\uparrow$ & Bleu-2 $\uparrow$ & Bleu-4 $\uparrow$ & Info. $\uparrow$ & ppl. $\downarrow$ \\
            \midrule
            
            \textbf{LLaVA} & 49.4 & 22.1 & 42.0 & 14.7 & 85.3 & 43.3 & 29.2 & 15.4 & 93.7 & 0.139 \\
            
            \hspace{3mm} + \emph{f.g.} unlearn. & 46.1 & 21.8 & 44.0 & 13.9 & 80.0 & 43.4 & 29.2 & 15.5 & 92.3 & 0.143 \\
            
            \hspace{3mm} + Sent. Loss & 37.8 & 19.7 & 40.0 & 12.7 & 70.0 & 45.9 & 31.7 & 17.5 & 91.8 & 0.087 \\        
            
            \hspace{3mm} + \textbf{\ours} & 34.4 & 14.7 & 21.0 & 6.5 & 86.7 & 48.0 & 33.5 & 18.9 & 92.1 & 0.101 \\
            
            \bottomrule
        \end{tabular}
    }
    \caption{Ablation results on LLaVA across different unlearning granularities.}
    \label{tab:ablation}
\end{table*}

\paragraph{Experimental Settings.} Adversarial configurations include: 
(1) \textbf{Relearning Attack}, where the unlearned model is fine-tuned on a subset of the original hallucination-inducing data to simulate memory recovery, monitoring the rebound across comprehensive evaluation metrics; 
(2) \textbf{LoRA Fine-tuning}, which examines the stability of the unlearning outcome by applying standard LoRA fine-tuning with approximately 10,000 samples from the original training dataset, testing if the hallucination suppression can be easily reactivated via parameter-efficient fine-tuning; and 
(3) \textbf{Adversarial Prompting}, which challenges the model with prompts that mandate exhaustive object listing to test its resistance against instruction-induced hallucinations. 
Prompt template for adversarial evaluation and additional implementation details are provided in Appendix~\ref{app:prompt} and Appendix~\ref{app:exp_detail}.

\begin{figure}[t] 
    \centering
    \begin{subfigure}[b]{1.0\linewidth} 
        \centering
        \includegraphics[width=0.9\linewidth]{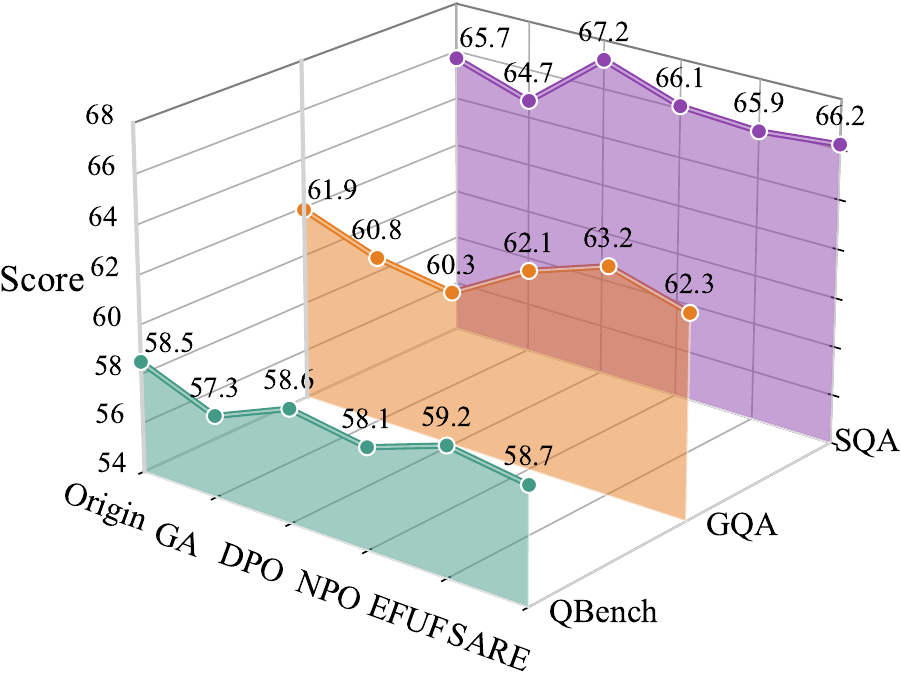}
        \caption{Evaluation results on GQA, SQA, and QBench.}
        \label{fig:mme}
    \end{subfigure}
    
    \vspace{0.5em} 
    
    \begin{subfigure}[b]{1.0\linewidth}
        \centering
        \includegraphics[width=0.9\linewidth]{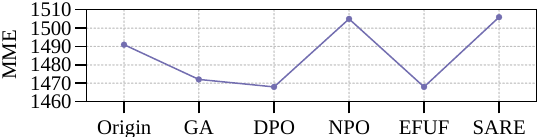}
        \caption{Scores on the MME benchmark.}
        \label{fig:general}
    \end{subfigure}

    \caption{Assessment of General Capabilities on GQA, SQA, QBench, and MME. \ours{} effectively maintains foundational reasoning and comprehension.}
    \label{fig:all_general_tasks}
\end{figure}

\subsection{Main Results}
\label{sec:main_results}

\paragraph{\textbf{RQ1: Can \ours{} effectively erase hallucinations and maintain robustness against parameter-based attacks?}}
Table~\ref{tab:main_exp} demonstrates the consistent superiority of \ours{} across diverse MLLM architectures. On mPLUG, our method achieves the most substantial reduction in hallucinations, slashing CHAIR$_s$ from the baseline 69.6 to 37.3, a significant improvement over EFUF's 43.6.
It also maintains superior grounding, evidenced by the highest POPE score on LLaVA. Beyond static evaluation, \ours{} exhibits exceptional stability against parameter-based attacks. While EFUF suffers from catastrophic memory resurgence as relearning data increases, \ours{} maintains a much flatter performance curve; specifically, under Relearn 140 on LLaVA, {\ours{}} limits the $\text{Human}_s$ rebound to 21.0 whereas EFUF surges to 29.0. Similarly, facing aggressive \textit{LoRA FT} perturbations, \ours{} suppresses CHAIR$_i$ to 17.4 on LLaVA, outperforming EFUF's 20.8. 
Such broad resilience validates that Targeted-SAM anchors the model in a flat loss region, robustly withstanding significant weight shifts.\footnote{Since computing exact sharpness metrics is computationally prohibitive for 7B-parameter MLLMs, we adopt the performance degradation under Relearning Attacks and LoRA Fine-tuning as a reliable empirical proxy for landscape sharpness.}

\begin{figure*}[ht]
    \centering
    \begin{minipage}{0.48\textwidth}
        \centering
        \includegraphics[width=0.83\linewidth]{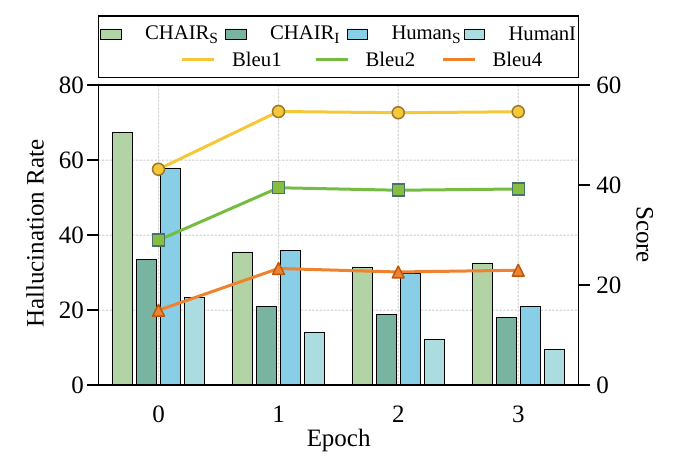}
        \caption{Training dynamics of \ours{}. Rapid convergence is achieved at Epoch 1, while further training leads to grounding collapse.}
        \label{fig:epoch}
    \end{minipage}
    \hfill 
    \begin{minipage}{0.48\textwidth}
        \centering
        \includegraphics[width=0.8\linewidth]{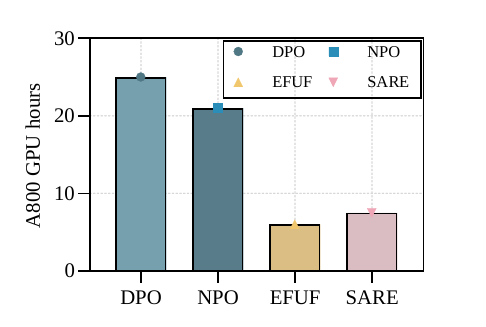}
        \caption{Efficiency comparison. \ours{} achieves significant speedup over DPO and NPO with competitive latency relative to EFUF.}
        \label{fig:time}
    \end{minipage}
\end{figure*}

\paragraph{\textbf{RQ2: Does \ours{} preserve general linguistic capabilities better than baselines?}} 
As presented in Table~\ref{tab:main_exp}, we evaluate the generation quality using BLEU, Perplexity, and Informativeness metrics. For semantic fidelity, \ours{} achieves a Bleu-4 of 18.9 on LLaVA, surpassing EFUF's 18.2 to demonstrate enhanced alignment without sacrificing coherence. Regarding fluency, \ours{} consistently outperforms EFUF by maintaining a PPL of 0.101 on LLaVA against the latter's 0.113. Such superior fluency suggests that our optimization avoids linguistic degradation while enhancing text smoothness. Finally, Informativeness results confirm the preservation of semantic richness; on mPLUG-Owl, \ours{} maintains a competitive 89.6, a value only 0.4 lower than EFUF, ensuring the model provides detailed descriptions instead of evasive responses.

\paragraph{\textbf{RQ3: Is the defense mechanism of \ours{} robust against perturbations?}} 
We further evaluate defense stability against input-level \textit{Adversarial Prompting}. As shown in Table~\ref{tab:main_exp}, EFUF exhibits significant vulnerability on LLaVA, where its CHAIR$_s$ spikes to 50.5, nearly equivalent to the baseline. In contrast, \ours{} demonstrates robust resistance to such perturbations. On mPLUG, the disparity becomes particularly distinct as \ours{} maintains a low CHAIR$_s$ of 37.5 while EFUF surges to 50.0. Such resilience confirms that our method transcends superficial pattern matching, fundamentally strengthening the model's reliance on genuine visual evidence rather than spurious correlations.

\subsection{Ablation Analysis}
\label{sec:ablation_results}

To investigate the individual contributions of each component in our framework, we conducted an ablation study on the LLaVA-v1.5-7B, comprehensively evaluating performance across both hallucination rates and generation quality. We compared four comparative settings: (1) \textbf{Origin}, representing the baseline LLaVA model; (2) \textbf{Fine-Grained Unlearning}, which exclusively employs the negative and positive subsentence datasets for targeted erasure and retention, excluding the sentence-level dataset; (3) \textbf{Sentence Loss Only}, which utilizes solely the sentence-level dataset for global consistency; and (4) \textbf{\ours}, the complete framework.

\paragraph{\textbf{Effects of Fine-Grained Unlearning.}}
As detailed in Table~\ref{tab:ablation}, applying fine-grained unlearning in isolation yields only marginal improvements in hallucination reduction. For instance, CHAIR$_s$ decreases trivially from 49.4 to 46.1. However, these limited gains come at the cost of linguistic coherence. The decline in fluency is highlighted by the increased PPL and reduced informativeness. This suggests that myopic token-level suppression, without broader contextual constraints, disrupts the pre-trained language manifold.

\paragraph{\textbf{Effects of Sentence Loss.}}
The Sentence Loss variant yields superior fluency with the lowest PPL of 0.087. While hallucination metrics appear to drop drastically, this improvement is deceptive. It correlates with a severe collapse in the POPE score from 85.3 to 70.0, indicating a failure in visual grounding. In this scenario, the model decouples visual inputs to generate safe but generic text, effectively sacrificing its fundamental visual recognition ability to satisfy the unlearning objective.


\subsection{Reasoning and Comprehension Analysis}

To holistically evaluate \ours{}, we measure its fine-grained reasoning, scientific understanding, and general perception across four benchmarks: MME~\cite{MME}, GQA~\cite{GQA}, ScienceQA~\cite{Science_QA}, and QBench~\cite{QBench}. 
Figure~\ref{fig:all_general_tasks} compares \ours{} against the baseline model and standard mitigation strategies, including GA~\cite{JangYYCLLS23}, DPO~\cite{DPO}, NPO~\cite{NPO}, and EFUF. \ours{} effectively maintains the model's foundational capabilities across diverse benchmarks. 
Notably, it achieves a top MME score of 1506 and remains highly competitive across GQA, ScienceQA, and QBench.
This confirms that \ours{} effectively mitigates hallucinations while maintaining robust reasoning capabilities.

\subsection{Training Dynamics Analysis}

We investigated training efficiency on mPLUG-Owl-7B. As illustrated in Figure~\ref{fig:epoch}, \ours{} exhibits rapid convergence, requiring only a single epoch to significantly reduce hallucinations and boost generation quality. Extending training beyond this point incurs unnecessary computational overhead while triggering a sharp POPE decline. This degradation renders the continued hallucination reduction meaningless, signaling a grounding collapse where the model sacrifices essential visual discrimination for superficial safety. Therefore, we select Epoch 1 as the optimal checkpoint, achieving a superior trade-off between erasure efficacy, visual grounding, and training costs.

\subsection{Efficiency Analysis}

We assessed the computational overhead of different methods using NVIDIA A800 GPUs. As reported in Figure~\ref{fig:time}, \ours{} demonstrates remarkable efficiency, achieving a significant speedup compared to computation-intensive baselines like DPO and NPO. While its latency is marginally higher than EFUF, \ours{} provides a superior trade-off by delivering substantially stronger hallucination mitigation with negligible additional cost. To explicitly quantify this marginal difference, we conducted a side-by-side wall-clock benchmark on LLaVA-v1.5-7B. Our measurements show that \ours{} exhibits a per-step slowdown of merely 1.07$\times$ compared to EFUF. This remarkable efficiency stems from the architectural characteristics of MLLM parameter-efficient fine-tuning. Although Targeted-SAM conceptually requires a dual-step optimization, the perturbed forward pass completely bypasses the computationally heavy vision encoder by reusing the cached visual embeddings. Furthermore, restricting the gradient computation solely to the lightweight safety expert minimizes the backward pass overhead. Consequently, the theoretical computational burden of SAM is largely absorbed, making \ours{} highly viable for robust and efficient real-world deployment.

\subsection{More Experiments}

To further substantiate the comprehensive effectiveness and broad applicability of \ours{}, we provide extensive supplementary analyses in the appendices. As presented in Appendix~\ref{app:generalizability}, we evaluate the generalizability of our framework across diverse MLLM architectures. Additionally, we explore the impact of varying core parameter configurations in Appendix~\ref{app:hyperparameter}. This includes detailed sensitivity analyses on the perturbation radius $\rho$, as well as the multi-objective loss weights $\lambda_1$ and $\lambda_2$, offering deeper insights into the optimization dynamics of our framework. Finally, qualitative case studies in Appendix~\ref{app:case_study} further demonstrate our robust resistance to relearning compared to standard unlearning.

\section{Conclusion}
In this paper, we reveal a critical robustness gap in MLLM hallucination unlearning, where standard methods converge to sharp local minima, leaving models vulnerable to rapid hallucination resurgence. Based on this insight, we introduce \ours, a sharpness-aware framework that enforces robust erasure through geometric regularization. By reformulating unlearning as a targeted min-max optimization, \ours{} simulates worst-case attacks during training to flatten the loss landscape, ensuring that hallucination suppression remains stable even under parameter shifts. Extensive experiments demonstrate that \ours{} effectively resists relearning and fine-tuning attacks while preserving foundational reasoning and linguistic capabilities.

\section*{Limitations}
The limitations of our work reside in three primary aspects. Firstly, the construction of the unlearning dataset relies on a static filtering pipeline that intrinsically focuses on known error patterns. Consequently, its coverage of latent hallucination triggers remains limited. These triggers are typically dormant in the pre-trained weights but may activate under novel or out-of-distribution prompts. Developing dynamic data curation strategies to enhance robustness constitutes a necessary future step. Secondly, our robust erasure mechanism is currently optimized for object existence hallucinations. The framework has yet to be extended to more nuanced hallucination types, such as incorrect object attributes or fallacious spatial positioning. Although our geometric flattening is theoretically adaptable to these complex cases, the primary bottleneck remains data curation. Since existing vision-language metrics fundamentally struggle with fine-grained semantics, developing advanced data construction strategies remains a critical prerequisite for future extensions. Finally, beyond these perception-level limitations, applying our geometric regularization to higher-order abstract reasoning hallucinations, in which errors stem from fallacious logical chains instead of simple object misidentification, remains a significant future challenge.

\section*{Ethics Statement}
Our research aims to enhance the reliability of Multimodal Large Language Models by mitigating hallucinations, which is a critical step toward ensuring the trustworthiness of AI systems in real-world applications. We identify and address three primary ethical considerations.

First, our methodology for data curation relies exclusively on publicly available datasets and automated alignment tools, ensuring that no private or sensitive user data is utilized during the unlearning process. Such an approach minimizes privacy risks while maintaining transparency in how hallucination patterns are identified and erased.

Second, we commit to releasing our code to the research community to foster collective defensive advancements and believe the net impact of this research is strongly positive, offering a practical and principled step toward more reliable and trustworthy AI systems.

Finally, during drafting and revision, we used AI assistants to help optimize writing and improve clarity. No content was generated unsupervised or without verification.

\section*{Acknowledgments}
 
This work was supported by the National Natural Science Foundation of China (Nos.62506166, U2441285),  the Natural Science Foundation
of Jiangsu Province (No.BK20251365), the China Postdoctoral Science Foundation (No. 2025M774283), the Scientific Research Starting Foundation of
Nanjing University of Aeronautics and Astronautics (No.1015-YAH24096), and the High Performance
Computing Platform of Nanjing University of Aeronautics and Astronautics.
This research is also sponsored by the DiDi GAIA Collaborative Research Funds (No. CCF-DiDi GAIA202507) and CAAI-MindSpore Open Fund (CAAIXSJLJJ 2025 MindSpore 01), developed on OpenI Community.

\bibliography{latex/custom}

\appendix
\clearpage

\section{Detailed Experimental Setups}
\label{app:exp_setup}

\subsection{Dataset}
\label{app:dataset}

We utilize the MSCOCO~2014 dataset~\cite{MSCOCO} as our primary source, following the data construction protocol established in EFUF~\cite{EFUF}. To ensure a rigorous evaluation, we randomly reserve 3,200 images that are strictly excluded from the training process, allocating 1,600 images for validation and another 1,600 images for testing. The remaining images serve as the candidate pool for constructing our unlearning dataset. Unlike traditional supervised fine-tuning, our approach does not utilize ground-truth captions during training. Since our approach necessitates only the raw images and their associated text queries, the official annotations provided by MSCOCO are used exclusively for evaluation purposes. From the candidate pool, approximately 30,000 training samples are curated via a CLIP-based filtering strategy with empirical thresholds ($T_0, T_1, T_2$). These samples are distributed among positive subsentences ($D_{pos}$), negative subsentences ($D_{neg}$), and sentence-level retention data ($D_{sent}$).

\subsection{Metrics}
\label{app:metrics}

\paragraph{Metrics on Hallucination Evaluation}
\label{app:metrics1}

To quantify the degree of hallucination, we employ three metrics: CHAIR(automated caption assessment), MHumanEval(human-verified judgment) and POPE(visual perception capabilities).

\begin{itemize}

    \item \textbf{CHAIR.} Caption Hallucination Assessment with Image Relevance (CHAIR)~\citep{rohrbach-etal-2018-object} is a widely used image captioning metric that identifies hallucinated objects by comparing the extracted objects with ground truth labels and evaluates both at the instance level (CHAIR$_I$) and sentence level (CHAIR$_S$). Belows are the detailed formation of these two metrics.

    \begin{equation}
        \text{CHAIR}_I = \frac{|\{\text{hallucinated objects}\}|}{|\{\text{all mentioned objects}\}|}
    \end{equation}
    
    \begin{equation}
        \text{CHAIR}_S = \frac{|\{\text{hallucinated captions}\}|}{|\{\text{total captions}\}|}
    \end{equation}
    
    where a "hallucinated caption" is defined as any response containing at least one object not present in the ground truth.

    \item \textbf{MHumanEval.}Given the open-ended generation capabilities of MLLMs, the standard CHAIR metric faces limitations as it relies on MSCOCO annotations which only cover a restricted set of pre-defined object categories, inevitably causing inaccuracies in evaluation. To address this, we incorporate human judgment into our evaluation. Following \cite{YuYZHHCHL0024}, we conduct a manual audit on a randomly sampled set of 100 generated captions. To ensure comparability with the CHAIR metric, we compute human-verified hallucination rates at both the instance and sentence granularities, providing a rigorous measurement of model reliability beyond the constraints of fixed vocabularies.

    \item \textbf{POPE.}The Polling-based Object Probing Evaluation (POPE)~\citep{POPE} is a VQA-based evaluation protocol designed to probe the model's visual perception stability. It queries the model with simple "Yes/No" questions regarding the existence of specific objects in the image. To prevent the model from exploiting statistical biases, POPE employs three sampling settings: Random (randomly sampling non-existent objects), Popular (sampling frequent but non-existent objects), and Adversarial (sampling co-occurring but non-existent objects). We report the F1 score for performance evaluation.

\end{itemize}

\paragraph{Metrics on Generation Quality Evaluation}
\label{app:metrics2}

While the primary objective of our method is to mitigate hallucinations, it is equally critical to ensure that the unlearning process does not degrade the model's general linguistic capabilities. We employ three distinct metrics to evaluate the consistency, utility and fluency of the generated content.

\begin{itemize}

    \item \textbf{BLEU.} To measure the lexical alignment between the generated captions and human-written references, we utilize the BLEU metric~\citep{BLEU}. BLEU calculates the precision of n-gram overlaps, serving as a standard proxy for linguistic consistency. While less sensitive to semantics than LLM-based metrics, it remains a vital benchmark for ensuring that the unlearning process does not catastrophically alter the model's vocabulary usage or sentence structure.

    \item \textbf{Informativeness.} Standard lexical metrics often fail to gauge whether key visual concepts are preserved. To address this, we implement a Semantic Coverage Score utilizing DeepSeek~\citep{deepseekv3} as an external judge. Instead of simple text matching, we prompt the evaluator to analyze the semantic alignment between the model's response and the ground-truth captions. This metric specifically quantifies the recall of essential visual details present in the reference, serving as a high-level proxy for model utility.

    \item \textbf{Perplexity.} To quantify the linguistic naturalness and coherence of the generated captions, we compute Perplexity (ppl.) using an external pre-trained GPT-2 model~\citep{GPT2}. Mathematically, this metric represents the exponentiated average negative log-likelihood of the generated token sequence. A lower PPL score indicates that the output is statistically closer to the distribution of natural human language, serving as a critical indicator that our unlearning intervention has not compromised the fundamental language modeling capabilities of the MLLM.
    
\end{itemize}

\subsection{Implementation Details}
\label{app:exp_detail}
We implement all models using the PyTorch framework~\cite{pytorch} and conduct experiments on an NVIDIA A800 GPU. During unlearning, we only tune the multimodal mapping layers of each MLLM to maintain architectural integrity. All models are trained for a fixed 1 epoch using the AdamW optimizer with a learning rate $\eta$ of 1e-5 and weight decay of 0.05. For our \ours~ framework, the unlearning loss weight $\lambda_1$ and sentence loss weight $\lambda_2$ are set to 0.3 and 0.3, respectively, while the perturbation radius $\rho$ is set to 0.05. Regarding the data curation pipeline described in Sec.~\ref{subsec:data_loss}, the thresholds for visual anchors ($T_0$) and hallucinated targets ($T_1$) are established at 32 and 23. Furthermore, to balance the sample distribution between sentence-level and subsentence-level data, the reliability threshold ($T_2$) for the sentence-level dataset $D_{sent}$ is set to 27.5.

\section{Derivation of Gradient Approximation}
\label{app:sam_derivation}

In this section, we provide the detailed mathematical justification for approximating the explicit Hessian computation with a second forward-backward pass. 
Since the retention losses are independent of the perturbation $\epsilon$, we focus our analysis exclusively on the hallucination component. In this context, the Targeted-SAM objective effectively minimizes a regularized loss defined as:
\begin{equation}
    \mathcal{J}_{neg}(\theta_{\phi}) = \mathcal{L}_{neg}(\theta_{\phi}) + \rho \|\nabla_{\theta_{\phi}} \mathcal{L}_{neg}(\theta_{\phi})\|_2.
\end{equation}
To update the parameters, we require the gradient of this objective with respect to $\theta_{\phi}$.

First, we analyze the derivative of the gradient norm regularization term $\rho \|\nabla_{\theta_{\phi}} \mathcal{L}_{neg}(\theta_{\phi})\|_2$. Let $g = \nabla_{\theta_{\phi}} \mathcal{L}_{neg}(\theta_{\phi})$ denote the gradient vector. The gradient of its $L_2$ norm is derived using the chain rule:
\begin{equation}
\begin{split}
    \nabla_{\theta_{\phi}} \|g\|_2 &= \nabla_{\theta_{\phi}} (g^\top g)^{1/2} \\
    &= \frac{1}{2(g^\top g)^{1/2}} \cdot \nabla_{\theta_{\phi}} (g^\top g) \\
    &= \frac{1}{2\|g\|_2} \cdot (2 \mathbf{H} g) \\
    &= \frac{\mathbf{H} \nabla_{\theta_{\phi}} \mathcal{L}_{neg}(\theta_{\phi})}{\|\nabla_{\theta_{\phi}} \mathcal{L}_{neg}(\theta_{\phi})\|_2},
\end{split}
\end{equation}
where $\mathbf{H} = \nabla^2_{\theta_{\phi}} \mathcal{L}_{neg}(\theta_{\phi})$ is the Hessian matrix, and we utilize the property $\nabla_{\theta_{\phi}}(g^\top g) = 2 \mathbf{H} g$. 

Substituting this result back into the gradient of the total objective $\mathcal{J}_{neg}$, the \textbf{theoretical gradient} is derived as:
\begin{equation}
\label{eq:true_grad}
\begin{split}
    \nabla_{\theta_{\phi}} \mathcal{J}_{neg} &= \nabla_{\theta_{\phi}} \mathcal{L}_{neg}(\theta_{\phi}) \\
    &\quad + \rho \frac{\mathbf{H} \nabla_{\theta_{\phi}} \mathcal{L}_{neg}(\theta_{\phi})}{\|\nabla_{\theta_{\phi}} \mathcal{L}_{neg}(\theta_{\phi})\|_2}.
\end{split}
\end{equation}
This confirms that the exact update direction explicitly involves the Hessian-vector product $\mathbf{H}v$, where $v$ is the normalized gradient direction.

Directly computing $\mathbf{H}v$ is computationally prohibitive. However, we demonstrate that this term naturally arises from the gradient at the perturbed state. Consider the gradient computed at $\theta_{\phi} + \epsilon^*$, where $\epsilon^* = \rho \frac{\nabla_{\theta_{\phi}} \mathcal{L}_{neg}(\theta_{\phi})}{\|\nabla_{\theta_{\phi}} \mathcal{L}_{neg}(\theta_{\phi})\|_2}$. We apply a first-order Taylor expansion to the gradient function $\nabla_{\theta_{\phi}} \mathcal{L}_{neg}(\cdot)$ around $\theta_{\phi}$:
\begin{equation}
\label{eq:approx_grad}
\begin{split}
    &\nabla_{\theta_{\phi}} \mathcal{L}_{neg}(\theta_{\phi} + \epsilon^*) \\
    &\approx \nabla_{\theta_{\phi}} \mathcal{L}_{neg}(\theta_{\phi}) + \mathbf{H} \epsilon^* \\
    &= \nabla_{\theta_{\phi}} \mathcal{L}_{neg}(\theta_{\phi}) + \mathbf{H} \left( \rho \frac{\nabla_{\theta_{\phi}} \mathcal{L}_{neg}(\theta_{\phi})}{\|\nabla_{\theta_{\phi}} \mathcal{L}_{neg}(\theta_{\phi})\|_2} \right) \\
    &= \nabla_{\theta_{\phi}} \mathcal{L}_{neg}(\theta_{\phi}) + \rho \frac{\mathbf{H} \nabla_{\theta_{\phi}} \mathcal{L}_{neg}(\theta_{\phi})}{\|\nabla_{\theta_{\phi}} \mathcal{L}_{neg}(\theta_{\phi})\|_2}.
\end{split}
\end{equation}
Comparing Eq.~\eqref{eq:true_grad} and Eq.~\eqref{eq:approx_grad}, we observe that the gradient computed at the perturbed state serves as a first-order approximation to the theoretical SAM gradient. 
This proves that our approach implicitly captures the curvature information required for robust optimization, effectively bypassing the computationally expensive materialization of the Hessian matrix.

\section{Prompt Templates for Adversarial Evaluation}
\label{app:prompt}

\begin{PromptBox}{Adversarial prompt}

    Please provide a comprehensive description of the image. Systematically list all the clearly identifiable objects you see, ensuring you do not overlook any distinct items.
    
\end{PromptBox}

\begin{table*}[t]
    \centering
    \small
    \setlength{\tabcolsep}{5pt} 
    \renewcommand{\arraystretch}{1.2}
    
    \begin{tabular}{cc | ccc | ccccc}
        \toprule
        \multirow{2}{*}{\textbf{Model}} & \multirow{2}{*}{\textbf{Method}} & \multicolumn{3}{c|}{\textbf{Hallucination Rate}} & \multicolumn{5}{c}{\textbf{Generation Quality}} \\
        \cmidrule(lr){3-5} \cmidrule(lr){6-10}
         & & CHAIR$_S$ $\downarrow$ & CHAIR$_I$ $\downarrow$ & POPE $\uparrow$ & Bleu-1 $\uparrow$ & Bleu-2 $\uparrow$ & Bleu-4 $\uparrow$ & Info. $\uparrow$ & PPL $\downarrow$ \\
        \midrule
        
        \multirow{3}{*}{MiniGPT-4} 
        & Original & 45.9 & 23.2 & 81.0 & 43.8 & 29.5 & 15.5 & 86.7 & 0.134 \\
        & EFUF     & 38.9 & 21.1 & 82.3 & 45.6 & 31.1 & 16.7 & 87.5 & 0.121 \\  
        & Ours & 35.4 & 18.8 & 81.3 & 48.1 & 32.8 & 17.2 & 85.1 & 0.110 \\
        \midrule

        \multirow{3}{*}{ShareGPT4V} 
        & Original & 46.8 & 22.3 & 87.8 & 43.3 & 29.2 & 15.4 & 89.6 & 0.157 \\
        & EFUF     & 36.9 & 18.4 & 88.1 & 46.9 & 32.5 & 18.1 & 91.1 & 0.159 \\
        & Ours & 34.7 & 14.9 & 91.0 & 47.9 & 33.4 & 18.8 & 92.0 & 0.112 \\
        
        \bottomrule
    \end{tabular}
    \caption{Performance comparison on generalizability. We compare the original models with the EFUF baseline and our proposed method on MiniGPT-4 and ShareGPT4V benchmarks. \textbf{Bold} denotes the best performance.}
    \label{tab:generalizability}
\end{table*}

\begin{table*}[t]
\centering
\resizebox{\textwidth}{!}{
\begin{tabular}{c c | ccccc | ccccc}
\toprule
\multirow{2}{*}{\textbf{Parameter}} & \multirow{2}{*}{\textbf{Value}} & \multicolumn{5}{c|}{\textbf{Hallucination Rate}} & \multicolumn{5}{c}{\textbf{Generation Quality}} \\
 & & $\text{Chair}_S \downarrow$ & $\text{Chair}_I \downarrow$ & $\text{Human}_S \downarrow$ & $\text{Human}_I \downarrow$ & POPE $\uparrow$ & Bleu1 $\uparrow$ & Bleu2 $\uparrow$ & Bleu4 $\uparrow$ & Info. $\uparrow$ & ppl. $\downarrow$ \\ 
\midrule
\multirow{4}{*}{$\lambda_1$} 
 & 0.1 & 49.0 & 21.9 & 40.0 & 14.1 & 87.9 & 43.6 & 29.5 & 15.6 & 92.8 & 0.125 \\
 & 0.2 & 30.2 & 14.6 & 18.0 & 6.3 & 73.4 & 47.6 & 33.1 & 18.6 & 92.3 & 0.100 \\
 & 0.3 & 29.1 & 13.6 & 18.0 & 6.1 & 85.9 & 47.7 & 33.2 & 18.7 & 92.3 & 0.101 \\
 & 0.4 & 50.4 & 22.4 & 42.0 & 14.0 & 89.3 & 43.6 & 29.4 & 15.6 & 92.6 & 0.126 \\
\cmidrule{1-12}

\multirow{4}{*}{$\lambda_2$} 
 & 0.1 & 50.4 & 22.4 & 42.0 & 13.7 & 90.0 & 43.5 & 29.4 & 15.5 & 92.9 & 0.125 \\
 & 0.2 & 46.1 & 20.2 & 38.0 & 11.3 & 79.7 & 44.6 & 30.3 & 16.3 & 92.5 & 0.110 \\
 & 0.3 & 29.1 & 13.6 & 18.0 & 6.1 & 85.9 & 47.7 & 33.2 & 18.7 & 92.3 & 0.101 \\
 & 0.4 & 43.8 & 21.7 & 35.0 & 12.7 & 79.6 & 43.7 & 29.6 & 15.7 & 92.6 & 0.115 \\ 
\cmidrule{1-12}

\multirow{4}{*}{$\rho$} 
 & 0.01 & 44.0 & 21.2 & 35.0 & 11.8 & 70.0 & 43.7 & 29.6 & 15.8 & 93.1 & 0.100 \\ 
 & 0.05 & 29.1 & 13.6 & 18.0 & 6.1 & 85.9 & 47.7 & 33.2 & 18.7 & 92.3 & 0.101 \\
 & 0.10 & 29.5 & 14.0 & 19.0 & 6.8 & 70.6 & 47.7 & 33.2 & 18.6 & 92.0 & 0.101 \\
 & 0.15 & 50.2 & 22.1 & 41.0 & 13.0 & 89.2 & 43.6 & 29.4 & 15.6 & 93.3 & 0.125 \\

\bottomrule
\end{tabular}
}
\caption{Hyperparameter Sensitivity Analysis on LLaVA-v1.5-7B. We investigate the impact of the negative loss weight $\lambda_1$, the sentence loss weight $\lambda_2$, and the perturbation radius $\rho$. The best configuration is highlighted in bold.}
\label{tab:hyperparameter_exp}
\end{table*}

\section{More Experiments}
\label{app:more_experimnets}

\subsection{Generalizability Analysis}
\label{app:generalizability}

To demonstrate the versatility and model-agnostic nature of our proposed framework, we extend our evaluation to two other representative Multimodal Large Language Models (MLLMs): MiniGPT-4~\cite{MiniGPT4EV} and ShareGPT4V~\cite{ShareGPT4VIL}. 
As summarized in Table~\ref{tab:generalizability}, our framework exhibits robust generalizability across diverse architectures, consistently surpassing both the original models and the EFUF baseline. 

Notably, results on MiniGPT-4 reveal a significant deviation from the typical trade-off between unlearning efficacy and model utility. While standard unlearning often compromises general capabilities, our method achieves superior linguistic fluency compared to the original model. This indicates that our targeted sharpness tuning operates with high precision: by flattening the loss landscape, it effectively prunes specific hallucination patterns and refines the model's probability distribution, rather than indiscriminately damaging its knowledge base.

This architectural robustness is further corroborated on ShareGPT4V, where \ours~ establishes comprehensive superiority across both safety and generation quality metrics. These results confirm that the erasure of hallucinations is not merely a superficial suppression, but a stable optimization that preserves, and in some instances enhances, the fundamental generative capabilities of the MLLM.

\subsection{Hyperparameter Sensitivity Analysis}
\label{app:hyperparameter}

In this segment, we delve into the effects of varying three critical hyperparameters: the negative loss weight $\lambda_1$, the sentence-level preservation weight $\lambda_2$, and the perturbation radius $\rho$. Our investigation, conducted on the LLaVA-v1.5-7B model, aims to understand how adjustments in these parameters influence the delicate trade-off between suppressing hallucinations and maintaining generation quality. The empirical results are summarized in Table~\ref{tab:hyperparameter_exp}.

\begin{itemize}
    \item \textbf{Impact of Negative Loss Weight $\lambda_1$.} 
    The coefficient $\lambda_1$ governs the magnitude of the penalty applied to hallucinated samples. As $\lambda_1$ increases from 0.1 to 0.3, we observe significant improvements in both hallucination reduction and generation quality, suggesting that a sufficient penalty is requisite to push the model out of hallucination-prone regions. However, a further increase to 0.4 triggers a sharp degradation across all metrics (e.g., PPL rises to 0.126). This performance drop likely stems from \textbf{distribution collapse}: an excessively aggressive unlearning penalty disrupts the model's linguistic manifold, forcing the probability mass to shift unpredictably. This destabilizes the optimization, causing both catastrophic forgetting of valid knowledge and a resurgence of hallucinations due to the broken probability distribution. Consequently, $\lambda_1 = 0.3$ is identified as the optimal balance point.

    \item \textbf{Impact of Sentence Loss Weight $\lambda_2$.} 
    The coefficient $\lambda_2$ regulates the importance of the sentence-level objective, serving as an anchor to preserve capabilities. While a moderate $\lambda_2$ (0.3) acts as a necessary stabilizer, increasing it to 0.4 results in a global performance decline---hallucination rates rise, and generation quality deteriorates (e.g., Bleu-1 drops and PPL worsens). We attribute this to \textbf{over-regularization and gradient conflict}. An overly dominant $\lambda_2$ imposes rigid constraints that conflict with the gradient updates required for unlearning ($L_{neg}$) and visual alignment ($L_{pos}$). This high tension prevents the model from converging to an optimal solution, trapping it in a suboptimal state where neither visual grounding nor linguistic fluency is effectively maintained. Thus, $\lambda_2 = 0.3$ is selected as the optimal setting.

    \item \textbf{Impact of Perturbation Radius $\rho$.} 
    The perturbation parameter $\rho$ controls the magnitude of the worst-case noise injected during optimization. When $\rho$ is small (0.01), the method yields limited gains, behaving similarly to standard fine-tuning. Increasing $\rho$ to 0.05 significantly enhances performance, confirming that an appropriate level of perturbation effectively flattens the loss landscape. Notably, performance degrades drastically when $\rho$ exceeds 0.10 (reaching 0.15). This drop suggests that an overly large perturbation radius pushes the model parameters too far from the optimal manifold, making it difficult for the outer minimization step to recover a valid solution. Based on these observations, we adopt $\rho = 0.05$ as the default configuration.
\end{itemize}

\section{Case Study}
\label{app:case_study}
To provide a more intuitive understanding of how \ours{} resists hallucination resurgence, we present qualitative comparisons between standard unlearning and our method under relearning attacks.

\begin{figure*}[t]  
    \centering  
    \includegraphics[width=0.99\linewidth]{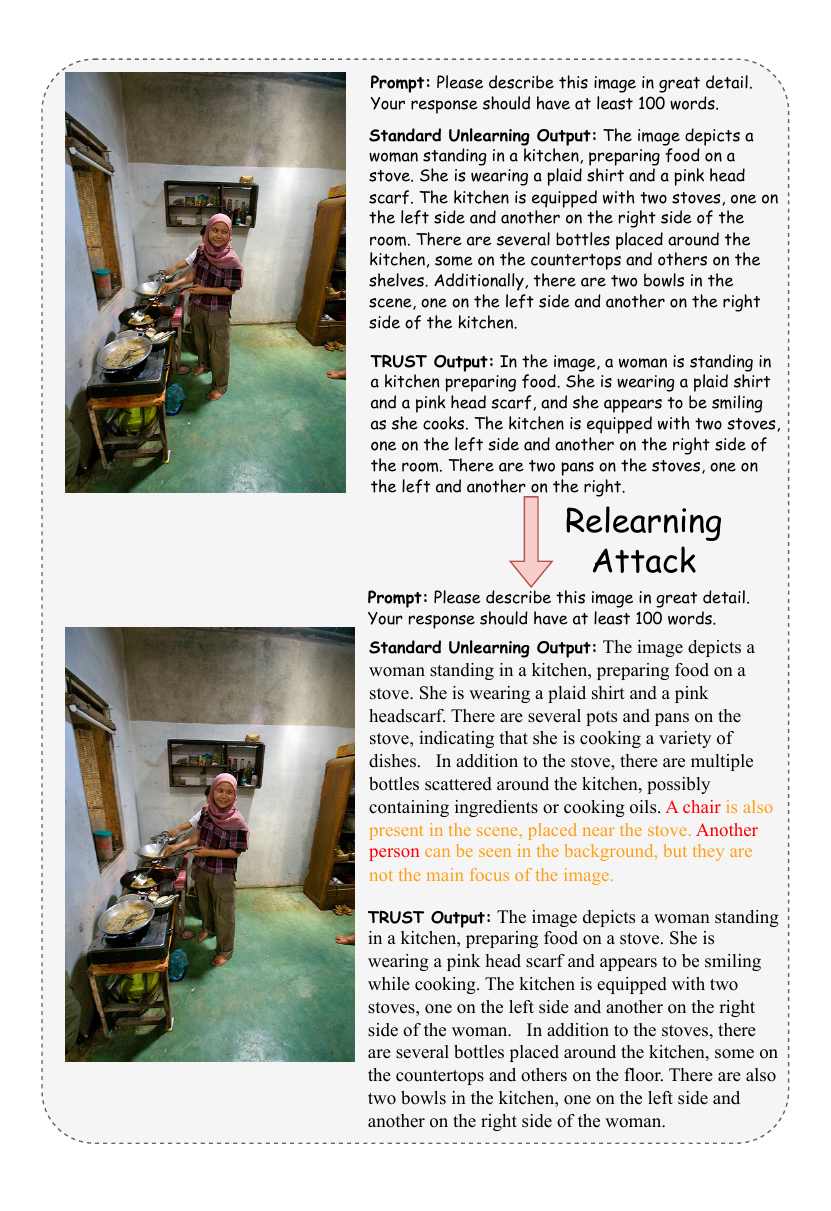}  
    \caption{Qualitative comparisons demonstrating the superior robustness of \ours{} against relearning attacks in contrast to standard unlearning. The red text denotes specific hallucinated content, while the orange indicates sentences containing hallucinations.} 
    \label{fig:case_study1}  
\end{figure*} 

\begin{figure*}[t]  
    \centering  
    \includegraphics[width=0.99\linewidth]{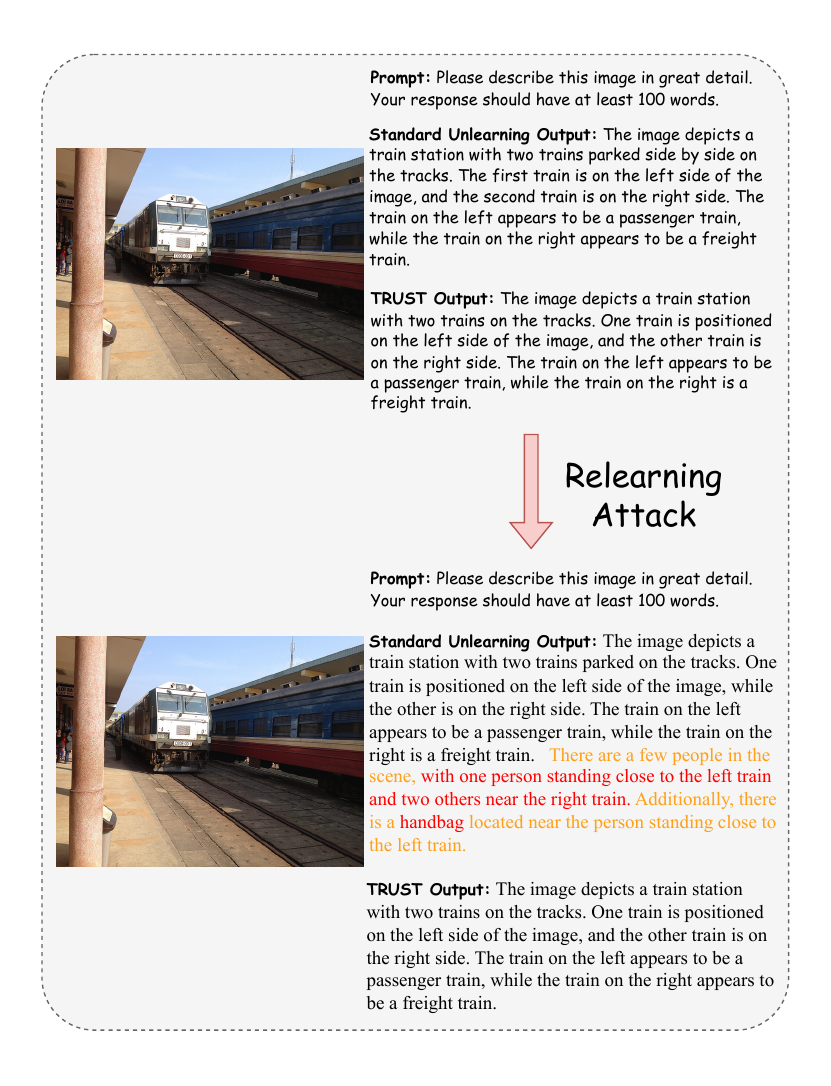}  
    \caption{Qualitative comparisons demonstrating the superior robustness of \ours{} against relearning attacks in contrast to standard unlearning. The red text denotes specific hallucinated content, while the orange indicates sentences containing hallucinations.} 
    \label{fig:case_study2}  
\end{figure*}

\end{document}